\documentclass{article}

\PassOptionsToPackage{numbers, compress}{natbib}
\usepackage[final]{neurips_2019}

\usepackage[utf8]{inputenc} % allow utf-8 input
\usepackage[T1]{fontenc}    % use 8-bit T1 fonts
\usepackage{hyperref}       % hyperlinks
\usepackage{url}            % simple URL typesetting
\usepackage{booktabs}       % professional-quality tables
\usepackage{amsfonts}       % blackboard math symbols
\usepackage{nicefrac}       % compact symbols for 1/2, etc.
\usepackage{microtype}      % microtypography

\usepackage[title]{appendix}

\usepackage{multirow}
\usepackage{amsmath}
\usepackage{capt-of}
\usepackage{tabularx}
\usepackage[caption=false]{subfig}
\usepackage{epsfig}
\usepackage{amssymb}
\usepackage{scalerel}
\usepackage[inline]{enumitem}
\usepackage{listings}
\usepackage{varwidth}
\usepackage[export]{adjustbox}
\usepackage{tikz}
\usetikzlibrary{tikzmark}
\usepackage{todonotes}
\usepackage{cleveref}

\usepackage{stmaryrd}
\usepackage{bbm}

\newcommand{\tabincell}[2]{\begin{tabular}{@{}#1@{}}#2\end{tabular}}

\newcommand{\sptk}[1]{\texttt{[#1]}}
\newcommand{\eqform}[1]{Equation~(\ref{#1})}
\newcommand{\newcite}[1]{\citeauthor{#1}~\citeyearpar{#1}}

\usepackage{algorithm}
\usepackage[noend]{algpseudocode}

\definecolor{deepblue}{rgb}{0,0,0.5}
\definecolor{officeblue}{RGB}{0,102,204}
\definecolor{deepred}{rgb}{0.6,0,0}
\definecolor{deepgreen}{rgb}{0,0.5,0}
\definecolor{mybrickred}{RGB}{182,50,28}

\definecolor{fillcolor}{RGB}{216,217,252}

\algnewcommand\algorithmicrequireb{{\hspace{0.95cm}}}
\algnewcommand\INPTDESCB{\item[\algorithmicrequireb]}

\algnewcommand\algorithmicfuncdesc{\textbf{Function:}}
\algnewcommand\FUNCDESC{\item[\algorithmicfuncdesc]}
\algnewcommand\algorithmicfuncdescb{{\hspace{0.86cm}}}
\algnewcommand\FUNCDESCB{\item[\algorithmicfuncdescb]}
\algnewcommand{\algorithmicgoto}{\textbf{goto}}
\algnewcommand{\Goto}[1]{\algorithmicgoto~\ref{#1}}

\usepackage{amsmath,amsfonts,bm}

\def\eqref#1{equation~\ref{#1}}
\def\1{\bm{1}}

\DeclareMathAlphabet{\mathsfit}{\encodingdefault}{\sfdefault}{m}{sl}
\SetMathAlphabet{\mathsfit}{bold}{\encodingdefault}{\sfdefault}{bx}{n}

\newcommand{\softmax}{\mathrm{softmax}}

\newcommand\ours{\textsc{UniLM}}

\title{Unified Language Model Pre-training for \\ Natural Language Understanding and Generation}

\author{{Li Dong\thanks{~~Equal contribution. $\dagger$ Contact person.} \quad Nan Yang\footnotemark[1] \quad Wenhui Wang\footnotemark[1] \quad Furu Wei\footnotemark[1] \ $^\dagger$ \quad Xiaodong Liu \quad Yu Wang} \\ {\bf Jianfeng Gao \quad Ming Zhou \quad Hsiao-Wuen Hon} \\
Microsoft Research \\
{\tt \{lidong1,nanya,wenwan,fuwei\}@microsoft.com} \\
{\tt \{xiaodl,yuwan,jfgao,mingzhou,hon\}@microsoft.com}}

\begin{document}

\maketitle

\begin{abstract}
This paper presents a new \textsc{\textbf{Uni}}fied pre-trained \textbf{L}anguage \textbf{M}odel (\ours{}) that can be fine-tuned for both natural language understanding and generation tasks. The model is pre-trained using three types of language modeling tasks: unidirectional, bidirectional, and sequence-to-sequence prediction. The unified modeling is achieved by employing a shared Transformer network and utilizing specific self-attention masks to control what context the prediction conditions on.
\ours{} compares favorably with BERT on the GLUE benchmark, and the SQuAD 2.0 and CoQA question answering tasks. 
Moreover, \ours{} achieves new state-of-the-art results on five natural language generation datasets, including improving the CNN/DailyMail abstractive summarization ROUGE-L to \textbf{40.51} ($2.04$ absolute improvement), the Gigaword abstractive summarization ROUGE-L to \textbf{35.75} ($0.86$ absolute improvement), the CoQA generative question answering F1 score to \textbf{82.5} ($37.1$ absolute improvement), the SQuAD question generation BLEU-4 to \textbf{22.12} ($3.75$ absolute improvement), and the DSTC7 document-grounded dialog response generation NIST-4 to \textbf{2.67} (human performance is 2.65). The code and pre-trained models are available at~\url{https://github.com/microsoft/unilm}.
\end{abstract}

\section{Introduction}

Language model (LM) pre-training has substantially advanced the state of the art across a variety of natural language processing tasks~\cite{semi_seq,elmo,ulmfit,gpt,bert,clozepretrain19}.
Pre-trained LMs learn contextualized text representations by predicting words based on their context using large amounts of text data, 
% After pre-training on large amounts of text data, 
and can be fine-tuned
% with additional task-specific layers 
to adapt to downstream tasks.
% various natural language understanding or generation tasks.

Different prediction tasks and training objectives have been used for pre-training LMs of different types, as shown in Table~\ref{tbl:compare:pretraining}.
ELMo~\cite{elmo} learns two unidirectional LMs: 
%based on long short-term memory networks~\cite{lstm}. 
a forward LM reads the text from left to right, and a backward LM encodes the text from right to left.
GPT~\cite{gpt} uses a left-to-right Transformer~\cite{transformer} to predict a text sequence word-by-word. In contrast, BERT~\cite{bert} employs a bidirectional Transformer encoder to fuse both the left and right context to predict the masked words.
% Moreover, BERT can explicitly model the relationship of a pair of texts, which has shown to be beneficial to many pair-wise natural language understanding tasks, such as natural language inference.
Although BERT significantly improves the performance of a wide range of natural language understanding tasks~\cite{bert}, its bidirectionality nature makes it difficult to be applied to natural language generation tasks~\cite{bert_mouth}.

\begin{table}[t]
\centering
\small
\begin{tabular}{l c c c c}
\toprule
                    & ELMo       & GPT        & BERT       & \ours       \\ \midrule
Left-to-Right LM       & \checkmark & \checkmark &            & \checkmark \\
Right-to-Left LM       & \checkmark &            &            & \checkmark \\
Bidirectional LM       &            &            & \checkmark & \checkmark \\
% Text-Pair Pre-training &            &            & \checkmark & \checkmark \\
Sequence-to-Sequence LM          &            &            &            & \checkmark \\ \bottomrule
\end{tabular}
\normalsize
\caption{Comparison between language model (LM) pre-training objectives.}
\label{tbl:compare:pretraining}
\end{table}

\begin{table*}[t]
\centering
\small
\begin{tabular}{l l l l}
\toprule
\textbf{\tabincell{c}{Backbone \\ Network}} & \textbf{\tabincell{c}{LM Objectives of \\ Unified Pre-training}} & \textbf{What Unified LM Learns} & \textbf{Example Downstream Tasks}      \\ \midrule
\multirow{6}{*}{\tabincell{l}{Transformer \\ with shared \\ parameters \\ for all LM \\ objectives}} & Bidirectional LM & Bidirectional encoding   & \tabincell{l}{GLUE benchmark \\ Extractive question answering}  \\ \cmidrule{3-3}
 & Unidirectional LM & Unidirectional decoding  & Long text generation \\ \cmidrule{3-3}
 & Sequence-to-Sequence LM & \tabincell{l}{Unidirectional decoding \\ conditioned on \\ bidirectional encoding}      & \tabincell{l}{Abstractive summarization \\ Question generation \\ Generative question answering}  \\ \bottomrule
\end{tabular}
\normalsize
\caption{The unified LM is jointly pre-trained by multiple language modeling objectives, sharing the same parameters. We fine-tune and evaluate the pre-trained unified LM on various datasets, including both language understanding and generation tasks.}
\label{tbl:objective:task}
\end{table*}

In this work we propose a new \textsc{\textbf{Uni}}fied pre-trained \textbf{L}anguage \textbf{M}odel (\ours) that can be applied to both natural language understanding (NLU) and natural language generation (NLG) tasks. 
\ours{} is a multi-layer Transformer network, jointly pre-trained on large amounts of text, optimized for three types of unsupervised language modeling objectives as shown in Table~\ref{tbl:objective:task}. % with shared parameters.
%As shown in Table~\ref{tbl:objective:task}, we pretrain a deep Transformer network on a large-scale corpus by employing several language modeling objectives. The parameters of Transformer and word embeddings are all shared.
In particular, we design a set of cloze tasks \cite{taylor1953cloze} 
%for language models in Table~\ref{tbl:objective:task}, 
where a masked word is predicted based on its context. These cloze tasks differ in how the context is defined. 
For a left-to-right unidirectional LM, the context of the masked word to be predicted consists of all the words on its left.     
For a right-to-left unidirectional LM, the context consists of all the words on the right.
For a bidirectional LM, the context consists of the words on both the right and the left~\cite{bert}. 
For a sequence-to-sequence LM, the context of the to-be-predicted word in the second (target) sequence consists of all the words in the first (source) sequence and the words on the its left in the target sequence. 

Similar to BERT, the pre-trained \ours{} can be fine-tuned (with additional task-specific layers if necessary) to adapt to various downstream tasks. But unlike BERT which is used mainly for NLU tasks, \ours{} can be configured, using different self-attention masks (Section~\ref{sec:method}), to aggregate context for different types of language models, and thus can be used for both NLU and NLG tasks. 

%By sharing a Transformer network as the backbone model,  we pretrain unidirectional (i.e., left-to-right, and right-to-left), bidirectional, and sequence-to-sequence LMs. The main differences between these LM objectives are what kind of context is given for each word to compute the contextualized representation. The left-to-right LM can only access the left context to predict the next word. In contrast, the bidirectional LM can fuse both the left and the right context for each word token. For the sequence-to-sequence LM, we explicitly model a pair of texts, where the first text is accessible to all the words, while each word token in the second text only has access to the left context.
%As the pre-training procedure learns different ways to aggregate the context, we can then fine-tune the unified model according to the objective of the downstream task, which supports both language understanding and generation problems.

The proposed \ours{} has three main advantages.
First, the unified pre-training procedure leads to a single Transformer LM that uses the shared parameters and architecture for different types of LMs, alleviating the need of separately training and hosting multiple LMs.
Second, the parameter sharing makes the learned text representations more general because they are jointly optimized for different language modeling objectives where context is utilized in different ways, mitigating overfitting to any single LM task.
Third, in addition to its application to NLU tasks, the use of \ours{} as a sequence-to-sequence LM (Section~\ref{sec:s2s:lm}), makes it a natural choice for NLG, such as abstractive summarization and question generation.

% We pre-trained \ours{} on a large corpus, and fine-tuned the pre-trained model on the tasks as described in Table~\ref{tbl:objective:task}.
Experimental results show that our model, used as a bidirectional encoder, compares favorably with BERT on the GLUE benchmark and two extractive question answering tasks (i.e., SQuAD 2.0 and CoQA). 
In addition, we demonstrate the effectiveness of \ours{} on five NLG datasets, where it is used as a sequence-to-sequence model, creating new state-of-the-art results on CNN/DailyMail and Gigaword abstractive summarization, SQuAD question generation, CoQA generative question answering, and DSTC7 dialog response generation.
% We also present text examples generated by \ours{} for a case study.

\section{Unified Language Model Pre-training}
\label{sec:method}

\begin{figure*}[t]
\centering
\includegraphics[width=0.94\textwidth]{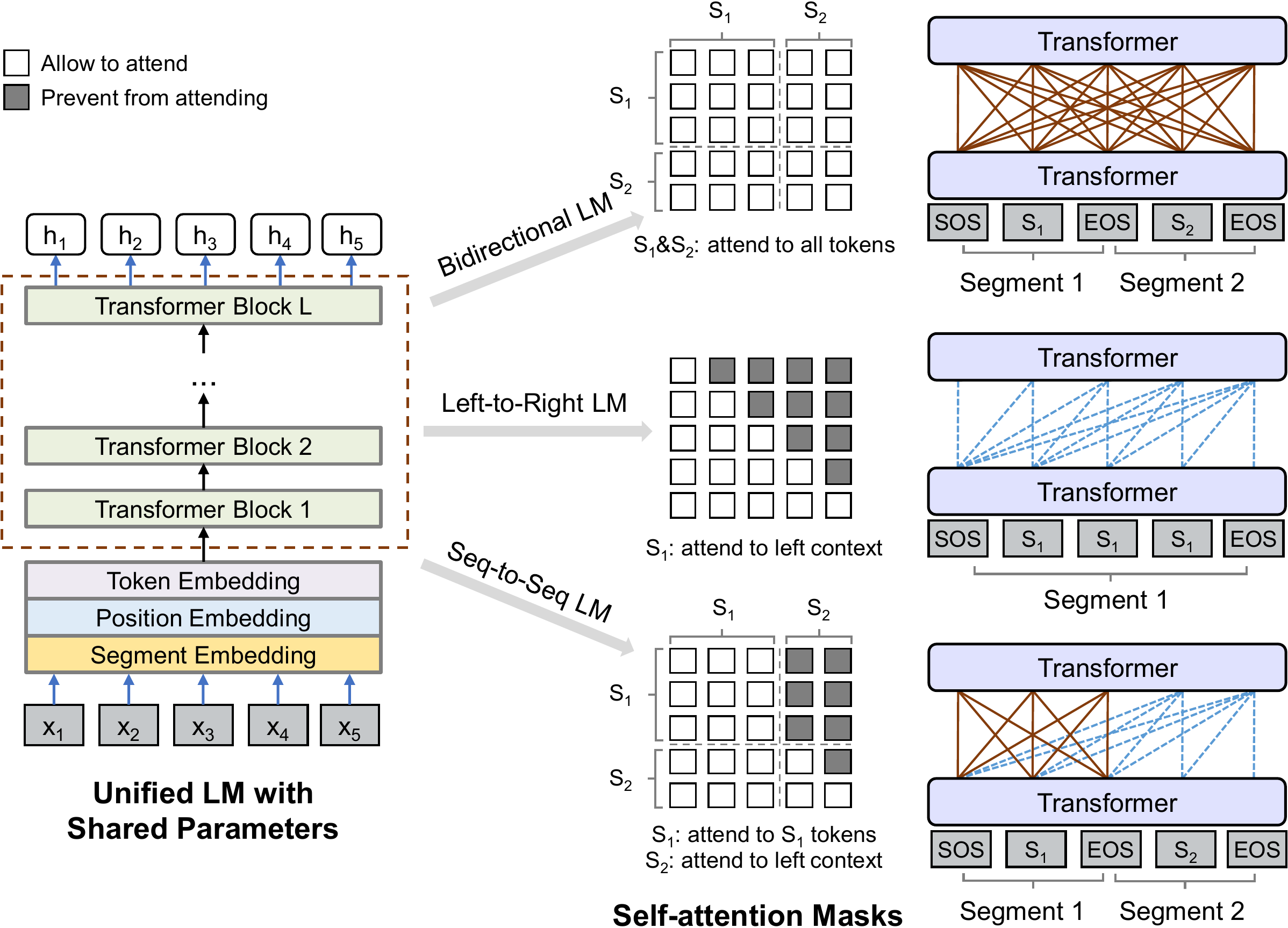}
\caption{Overview of unified LM pre-training. The model parameters are shared across the LM objectives (i.e., bidirectional LM, unidirectional LM, and sequence-to-sequence LM). We use different self-attention masks to control the access to context for each word token. The right-to-left LM is similar to the left-to-right one, which is omitted in the figure for brevity.}
\label{fig:method}
\end{figure*}

% \ours{} is based on a multi-layer Transformer network~\cite{transformer}.
Given an input sequence $x = x_{1} \cdots x_{|x|}$, \ours{} obtains a contextualized vector representation for each token.
%The input tokens are represented according to the word, the position, and the text segment it belongs to. Next, the input vectors are fed into a stack of multi-layer Transformer blocks, which uses self-attention to compute the text representations by considering the whole input sequence. 
%The vector representations are pre-trained on large amounts of text, optimized jointly with respect to a set of unsupervised language modeling objectives.
% in an unsupervised manner.
As shown in Figure~\ref{fig:method}, the pre-training optimizes the shared Transformer~\cite{transformer} network with respect to several unsupervised language modeling objectives, namely, unidirectional LM, bidirectional LM, and sequence-to-sequence LM.
In order to control the access to the context of the word token to be predicted, we employ different masks for self-attention. In other words, we use masking to control how much context the token should attend to when computing its contextualized representation.
Once \ours{} is pretrained, we can fine-tune it using task-specific data for downstream tasks.
%both natural language understanding and generation tasks.

\subsection{Input Representation}
\label{sec:input}

The input $x$ is a word sequence, which is either a text segment for unidirectional LMs or a pair of segments packed together for bidirectional LM and sequence-to-sequence LM. 
%There are two segments of sentences (i.e., a pair of texts) for both bidirectional LM and sequence-to-sequence LM objectives, while the unidirectional LM objective only contains one segment.
We always add a special start-of-sequence (\sptk{SOS}) token at the beginning of input, and a special end-of-sequence (\sptk{EOS}) token at the end of each segment. 
%The corresponding output vector can be used as the representation of the whole input.
% Moreover, we append a special end-of-sequence (\sptk{EOS}) token to the end of each segment. The token indicates the boundary of a pair of segments. 
\sptk{EOS} not only marks the sentence boundary in NLU tasks, but also is used for the model to learn when to terminate the decoding process in NLG tasks.
The input representation follows that of BERT~\cite{bert}.
Texts are tokenized to subword units by WordPiece~\cite{gnmt}.
%For instance, the word \tx{forecasted} is split to \tx{forecast} and \tx{\#\#ed}, where \tx{\#\#} indicates the the pieces are belong to one word.
For each input token, its vector representation is computed by summing the corresponding token embedding, position embedding, and segment embedding.
% We use absolute position embedding, which assigns different vectors for positions.
%In addition, we employ segment embeddings to differentiate a pair of text. To be specific, the first and the second segments are assigned with two different segment embeddings.
Since \ours{} is trained using multiple LM tasks, segment embeddings also play a role of LM identifier in that we use different segment embeddings for different LM objectives. 
% so that they can play a role of LM identifier.

\subsection{Backbone Network: Multi-Layer Transformer}

%We use a deep Transformer consisting of stacked self-attention layers~\cite{transformer} as the backbone network to encode contextual information.
%, consisting of stacked self-attention layers.
The input vectors $\{\textbf{x}_i\}_{i=1}^{|x|}$ is
%of input sequence, 
first packed into $\mathbf{H}^0 = [\mathbf{x}_1, \cdots, \mathbf{x}_{|x|}]$, and then
encoded into contextual representations at different levels of abstract
$\mathbf{H}^l = [\mathbf{h}_1^l, \cdots, \mathbf{h}_{|x|}^l]$ using an $L$-layer Transformer $\mathbf{H}^l = \mathrm{Transformer}_{l}(\mathbf{H}^{l-1}), l \in [1,L]$.
%where $l \in [1, L]$, and $\mathbf{H}^L = [\mathbf{h}_1^L, \cdots, \mathbf{h}_{|x|}^L]$.
%We use the hidden vector $\mathbf{h}_i^L$ as the contextualized representation of the input token $x_i$.
% \paragraph{Self-attention Masks}
In each Transformer block, multiple self-attention heads are used to aggregate the output vectors of the previous layer.
%The results of the attention heads are concatenated and linearly projected to form the output.
For the $l$-th Transformer layer, the output of a self-attention head  $\mathbf{A}_l$ is computed via:
\begin{align}
\mathbf{Q} &= \mathbf{H}^{l-1} \mathbf{W}_l^Q ,\quad \mathbf{K} = \mathbf{H}^{l-1} \mathbf{W}_l^K ,\quad \mathbf{V} = \mathbf{H}^{l-1} \mathbf{W}_l^V \\
\mathbf{M}_{ij} &= \begin{cases} 0, &\text{allow to attend} \\ -\infty, &\text{prevent from attending} \end{cases} \label{eq:att:mask} \\
\mathbf{A}_l &= \softmax(\frac{\mathbf{Q} \mathbf{K}^{\intercal}}{ \sqrt{d_k}} + \mathbf{M}) \mathbf{V}_l
\end{align}
where the previous layer's output $\mathbf{H}^{l-1} \in \mathbb{R}^{|x| \times d_h}$ is linearly projected to a triple of queries, keys and values using parameter matrices $\mathbf{W}_l^Q , \mathbf{W}_l^K , \mathbf{W}_l^V \in \mathbb{R}^{d_h \times d_k}$, respectively, and the mask matrix $\mathbf{M} \in \mathbb{R}^{|x| \times |x|} $ determines whether a pair of tokens can be attended to each other.

We use different mask matrices $\mathbf{M}$ to control what context a token can attend to when computing its contextualized representation, as illustrated in Figure~\ref{fig:method}. Take bidirectional LM as an example. The elements of the mask matrix are all $0$s, indicating that all the tokens have access to each other.
% For more implementation details of Transformer, we refer readers to~\citet{transformer}.

\subsection{Pre-training Objectives}
\label{sec:pretrain}

% We employ multiple LM objectives to pre-train \ours{} in an unsupervised manner. The main difference among these LMs is what context they encode for each word token. This is implemented using different self-attention masks as described in \eqform{eq:att:mask}.

We pretrain \ours{} using four cloze tasks designed for different language modeling objectives.
% They differ in what context to use in prediction. 
In a cloze task, 
%we mask some percentage of input tokens at random, and predict only those masked tokens using \ours.
%Specifically, 
we randomly choose some WordPiece tokens in the input, and replace them with special token \sptk{MASK}. 
Then, we feed their corresponding output vectors computed by the Transformer network into a softmax classifier to predict  the masked token. 
The parameters of \ours{} are learned to minimize the cross-entropy loss computed using the predicted tokens and the original tokens.
It is worth noting that the use of cloze tasks makes it possible to use the same training procedure for all LMs, unidirectional and bidirectional alike.
% In what follows, we detail the design of the four cloze tasks.
% The cloze task has been proven to be effective for pre-training in BERT~\cite{bert}.

\paragraph{Unidirectional LM}
\label{sec:uni:lm}

We use both left-to-right and right-to-left LM objectives.
%in the pre-training procedure. 
% During unidirectional LM pre-training, we use one segment (i.e., a span of contiguous text) for the input.
Take the left-to-right LM as an example. 
The representation of each token encodes %considers
only the leftward context tokens and itself.
% without depending on the future positions.
For instance, to predict the masked token of ``$x_1 x_2$ \sptk{MASK} $x_4$'', only tokens $x_1, x_2$ and itself can be used.
This is done by using a triangular matrix for the self-attention mask $\mathbf{M}$ (as in \eqform{eq:att:mask}), 
where the upper triangular part of the self-attention mask is set to $-\infty$, and the other elements to $0$, as shown in Figure~\ref{fig:method}.
%As shown in Figure~\ref{fig:method}, we set the upper triangular part of the self-attention mask to $-\infty$, and the other elements to $0$, which allows tokens to only attend to their earlier (left) positions.
Similarly, a right-to-left LM predicts a token conditioned on its future (right) context.

%Compared with standard unidirectional LMs (such as ELMo, and GPT), \ours{} uses cloze tasks for pre-training,
% does not need an explicit inference process in training,
%thus there is no position shift between input tokens and predictions. In other words, the conventional left-to-right LM predicts the next token by feeding the previous tokens (i.e., either the ground-truth tokens or the ones sampled by the model), while the proposed LM uses the \sptk{MASK} symbol to emit a predicted token.
%In terms of input and output token positions, the modification makes it possible to use the same training procedure for all LMs, unidirectional and bidirectional alike.

%makes unidirectional LMs aligned with the bidirectional LM objective.

\paragraph{Bidirectional LM}
\label{sec:bi:lm}

Following~\cite{bert}, a bidirectional LM allows all tokens to attend to each other in prediction. %during encoding in Transformer. 
It encodes contextual information from both directions, and can generate better contextual representations of text than its unidirectional counterpart.
As indicated in \eqform{eq:att:mask}, the self-attention mask $\mathbf{M}$ is a zero matrix, so that every token is allowed to attend across all positions in the input sequence.

\paragraph{Sequence-to-Sequence LM}
\label{sec:s2s:lm}

%A sequence-to-sequence LM takes two text segments as input. %The pair of sampled texts are contiguous in the corpus.
As shown in Figure~\ref{fig:method}, for prediction, the tokens in the first (source) segment can attend to each other from both directions within the segment, while
%But The tokens of the second (target) segment are blocked from attending.
%In contrast, 
the tokens of the second (target) segment can only attend to the leftward context in the target segment and itself, as well as all the tokens in the source segment.
For example, given source segment $t_1 t_2$ and its target segment $t_3 t_4 t_5$, we feed input ``\sptk{SOS} $t_1~t_2$ \sptk{EOS} $t_3~t_4~t_5$ \sptk{EOS}'' into the model.
While both $t_1$ and $t_2$ have access to the first four tokens, including \sptk{SOS} and \sptk{EOS}, $t_4$ can only attend to the first six tokens.
% , but neither $t_5$ nor the last \sptk{EOS}.

Figure~\ref{fig:method} shows the self-attention mask $\mathbf{M}$ used for the sequence-to-sequence LM objective. 
% To be specific, 
The left part of $\mathbf{M}$ is set to $0$ so that all tokens can attend to the first segment. The upper right part is set to $-\infty$ to block attentions from the source segment to the target segment. Moreover, for the lower right part, we set its upper triangular part to $-\infty$, and the other elements to $0$, which prevents tokens in the target segment from attending their future (right) positions.

During training, we randomly choose tokens in both segments, and replace them with the special token \sptk{MASK}. The model is learned to recover the masked tokens.
Since the pair of source and target texts are packed as a  contiguous input text sequence in training, we implicitly encourage the model to learn the relationship between the two segments.
In order to better predict tokens in the target segment, \ours{} learns to effectively encode the source segment.
Thus, the cloze task designed for the sequence-to-sequence LM, also known as the encoder-decoder model, simultaneously pre-trains a bidirectional encoder and an unidirectional decoder.
The pre-trained model, used as an encoder-decoder model, can be easily adapted to a wide range of conditional text generation tasks, such as abstractive summarization.
% , where the target segment is decoded (generated) by conditioning on the encoded vector of the source sentence

\paragraph{Next Sentence Prediction}
\label{sec:nsp}

% In addition to the LM objectives
For the bidirectional LM, we also include the next sentence prediction task for pre-training, as in~\cite{bert}.

%We take a pair of segments (\texttt{S1}, \texttt{S2}) as input, and predict whether \texttt{S2} is the next segment that follows \texttt{S1}. To be specific, for the segment \texttt{S1},~$50\%$ of the time we choose the text span that follows \texttt{S1} as the second segment \texttt{S2}, and~$50\%$ of the time a span randomly sampled from the corpus is used as \texttt{S2}. As described in Section~\ref{sec:input}, the packed input is ``\sptk{SOS} \texttt{S1} \sptk{EOS} \texttt{S2} \sptk{EOS}''. We then feed the encoding vector of \sptk{SOS} into a softmax classifier to make the prediction.

\subsection{Pre-training Setup}

The overall training objective the sum of different types of LM objectives described above.
Specifically, within one training batch, $1/3$ of the time we use the bidirectional LM objective, $1/3$ of the time we employ the sequence-to-sequence LM objective, and both left-to-right and right-to-left LM objectives are sampled with rate of $1/6$.
%For the bidirectional LM examples, we also added the mean likelihood of next sentence prediction.
The model architecture of \ours{} follows that of BERT$_\text{LARGE}$~\cite{bert} for a fair comparison. 
The gelu activation~\cite{gelu} is used as GPT~\cite{gpt}.
%We followed the similar model size and pre-training settings as BERT$_\text{LARGE}$~\cite{bert} for comparison purposes.
Specifically, we use a $24$-layer Transformer with $1,024$ hidden size, and $16$ attention heads, which contains about $340$M parameters. The weight matrix of the softmax classifier is tied with token embeddings.
\ours{} is initialized by BERT$_\text{LARGE}$, and then pre-trained using English Wikipedia\footnote{Wikipedia version: enwiki-20181101.} and BookCorpus~\cite{bookcorpus}, which have been processed in the same way as~\cite{bert}.
%following the preprocess and the WordPiece tokenization of~\citet{bert}.
The vocabulary size is $28,996$. The maximum length of input sequence is $512$. The token masking probability is $15\%$. 
Among masked positions, $80\%$ of the time we replace the token with \sptk{MASK}, $10\%$ of the time with a random token, and keeping the original token for the rest. In addition, $80\%$ of the time we randomly mask one token each time, and $20\%$ of the time we mask a bigram or a trigram.

Adam~\cite{adam} with $\beta_1=0.9$, $\beta_2=0.999$ is used for optimization. The learning rate is 3e-5, with linear warmup over the first $40,000$ steps and linear decay. The dropout rate is $0.1$. The weight decay is $0.01$. The batch size is $330$. 
The pre-training procedure runs for about $770,000$ steps. It takes about $7$ hours for $10,000$ steps using $8$ Nvidia Telsa V100 32GB GPU cards with mixed precision training.
% \footnote{More pre-training steps tend to yield better performance on downstream tasks.}

\subsection{Fine-tuning on Downstream NLU and NLG Tasks}
\label{sec:fine-tuning}

%We fine-tune the pre-trained \ours{} (with additional task-specific layers if necessary) for various downstream tasks. Compared to previous pre-trained LMs (such as GPT, and BERT), we use different self-attention masks (as in \eqform{eq:att:mask}) to adapt the same pre-trained model to both natural language understanding and generation tasks.

% Unlike previous pre-trained LMs (such as GPT, and BERT), which can only apply to either natural language understanding or generation tasks, we can use different self-attention masks, as in \eqform{eq:att:mask}, to fine-tune \ours{} to both natural language understanding and generation tasks.

For NLU tasks, we fine-tune \ours{} as a bidirectional Transformer encoder, like BERT. Take text classification as an example. We use the encoding vector of \sptk{SOS} as the representation of input, denoted as $\mathbf{h}_1^L$, and
feed it to a randomly initialized softmax classifier (i.e., the task-specific output layer), where 
the class probabilities are computed as $\softmax( \mathbf{h}_1^L \mathbf{W}^C )$, where $\mathbf{W}^C \in \mathbb{R}^{d_h \times C}$ is a parameter matrix, and $C$ the number of categories. 
We maximize the likelihood of the labeled training data by updating the parameters of the pre-trained LM and the added softmax classifier. 

For NLG tasks, we take the sequence-to-sequence task as an example. The fine-tuning procedure is similar to pre-training using the self-attention masks as in Section~\ref{sec:s2s:lm}.
Let \texttt{S1} and \texttt{S2} denote source and target sequences, respectively. 
We pack them together with special tokens, to form the input ``\sptk{SOS} \texttt{S1} \sptk{EOS} \texttt{S2} \sptk{EOS}''.
The model is fine-tuned by masking some percentage of tokens in the target sequence at random, and learning to recover the masked words. 
The training objective is to maximize the likelihood of masked tokens given context.
It is worth noting that \sptk{EOS}, which marks the end of the target sequence, can also be masked during fine-tuning, thus when this happens, the model learns when to emit \sptk{EOS} to terminate the generation process of the target sequence.

\section{Experiments}
\label{sec:exp}

We have conducted experiments on both NLU (i.e., the GLUE benchmark, and extractive question answering) and NLG tasks (i.e., abstractive summarization, question generation, generative question answering, and dialog response generation). 
% We also present text samples generated from \ours, which is used as a left-to-right unidirectional LM, for a case study.

\subsection{Abstractive Summarization}

Automatic text summarization produces a concise and fluent summary conveying the key information in the input (e.g., a news article).
We focus on abstractive summarization, a generation task where the summary is not constrained to reusing the phrases or sentences in the input text.
We use the non-anonymized version of the CNN/DailyMail dataset~\cite{see-etal-2017-get} and Gigaword~\cite{abs} for model fine-tuning and evaluation.
We fine-tune \ours{} as a sequence-to-sequence model following the procedure described in Section~\ref{sec:fine-tuning} by 
concatenating document (the first segment) and summary (the second segment) as input which is truncated according to a pre-defined maximum length. 
% We randomly replace the words in the summary with \sptk{MASK} to generate training samples.
% In addition, we use extractive summarization as an auxiliary training task. 
% We identify the approximate extractive oracle for training using the method described in~\cite{zhou-etal-2018-neural-document}.
% For each input sentence, we feed the output vector of the first token into a classifier, and predict whether the sentence appears in the extractive oracle.

We fine-tune our model on the training set for $30$ epochs.
We reuse most hyper-parameters from pre-training. The masking probability is $0.7$. We also use label smoothing~\cite{label:smoothing} with rate of $0.1$.
For CNN/DailyMail, we set batch size to $32$, and maximum length to $768$.
For Gigaword, we set batch size to $64$, and maximum length to $256$.
During decoding, we use beam search with beam size of $5$. The input document is truncated to the first $640$ and $192$ tokens for CNN/DailyMail and Gigaword, respectively.
We remove duplicated trigrams in beam search, and tweak the maximum summary length on the development set~\cite{Paulus2018ADR, Edunov2019PretrainedLM}.

\begin{table}[t]
\centering
\small
\begin{minipage}{2.7in}
\centering
\begin{tabular}{@{\hskip6pt}l@{\hskip6pt} @{\hskip6pt}l@{\hskip6pt} @{\hskip6pt}l@{\hskip6pt} @{\hskip6pt}l@{\hskip6pt}}
\toprule
                & RG-1       & RG-2        & RG-L  
                \\ \midrule
\multicolumn{4}{l}{\emph{Extractive Summarization}} \\
\quad LEAD-3 & 40.42           & 17.62   & 36.67 \\
\quad Best Extractive~\cite{bertsum} & 43.25 & \textbf{20.24} & 39.63 \\
\midrule
\multicolumn{4}{l}{\emph{Abstractive Summarization}} \\
\quad PGNet~\cite{see-etal-2017-get}   & 39.53 & 17.28 & 37.98  \\
\quad Bottom-Up~\cite{gehrmann-etal-2018-bottom}   & 41.22            & 18.68           & 38.34  \\ 
\quad S2S-ELMo~\cite{Edunov2019PretrainedLM}   & 41.56 & 18.94           &   38.47   \\
% \quad \ours & \textbf{43.47}  & \textbf{20.30} & \textbf{40.63} \\
\quad \ours & \textbf{43.33} & 20.21 & \textbf{40.51} \\
\bottomrule
\end{tabular}
\normalsize
\caption{Evaluation results on CNN/DailyMail summarization. Models in the first block are extractive systems listed here for reference, while the others are abstractive models. The results of the best reported extractive model are taken from~\cite{bertsum}. RG is short for ROUGE.}
\label{tbl:summary:result}
\end{minipage}
\hfill
\begin{minipage}{2.7in}
\centering
\begin{tabular}{@{\hskip6pt}l@{\hskip6pt} @{\hskip6pt}l@{\hskip6pt} @{\hskip6pt}l@{\hskip6pt} @{\hskip6pt}l@{\hskip6pt}}
\toprule
                & RG-1       & RG-2        & RG-L  
                \\ \midrule
\multicolumn{4}{l}{\emph{10K Training Examples}} \\
\quad Transformer~\cite{transformer}   & 10.97 & 2.23 & 10.42 \\
\quad MASS~\cite{mass} & 25.03 & 9.48 & 23.48 \\
\quad \ours & \textbf{32.96} & \textbf{14.68} & \textbf{30.56} \\
\midrule
\multicolumn{4}{l}{\emph{Full Training Set}} \\
\quad OpenNMT~\cite{opennmt}   & 36.73 & 17.86 & 33.68  \\
\quad Re3Sum~\cite{re3sum}   & 37.04 & 19.03 & 34.46 \\
\quad MASS~\cite{mass}   & 37.66 & 18.53 & 34.89 \\
\quad \ours & \textbf{38.45} & \textbf{19.45} & \textbf{35.75} \\
\bottomrule
\end{tabular}
\normalsize
\caption{Results on Gigaword abstractive summarization. Models in the first block only use 10K examples for training, while the others use 3.8M examples. Results of OpenNMT and Transformer are taken from~\cite{re3sum,mass}. RG is short for ROUGE.}
\label{tbl:summary:gigaword}
\end{minipage}
\end{table}

We use the F1 version of ROUGE~\cite{lin-2004-rouge} as the evaluation metric for both datasets.
In Table~\ref{tbl:summary:result}, we compare \ours{} against the baseline and several state-of-the-art models on CNN/DailyMail.
LEAD-3 is a baseline model that extracts the first three sentences in a document as its summary.
PGNet~\cite{see-etal-2017-get} is a sequence-to-sequence model based on the pointer-generator network.
S2S-ELMo~\cite{Edunov2019PretrainedLM} uses a sequence-to-sequence model augmented with pre-trained ELMo representations, which is termed as SRC-ELMO+SHDEMB in~\cite{Edunov2019PretrainedLM}.
Bottom-Up~\cite{gehrmann-etal-2018-bottom} is a sequence-to-sequence model augmented with a bottom-up content selector for selecting salient phrases.
We also include in Table~\ref{tbl:summary:result} the best reported extractive summarization result~\cite{bertsum} on the dataset.
% Following~\cite{see-etal-2017-get,gehrmann-etal-2018-bottom}, we use the F1 version of ROUGE~\cite{lin-2004-rouge} as the evaluation metric.
As shown in Table~\ref{tbl:summary:result}, our model outperforms all previous abstractive systems, creating a new state-of-the-art abstractive summarization result on the dataset. Our model also outperforms the best extractive model~\cite{bertsum} by $0.88$ point in ROUGE-L.

In Table~\ref{tbl:summary:gigaword}, we evaluate the models on Gigaword with different scales (10K and 3.8M).
Both Transformer~\cite{transformer} and OpenNMT~\cite{opennmt} implement standard attentional sequence-to-sequence models.
Re3Sum~\cite{re3sum} retrieves summaries as candidate templates, and then use an extended sequence-to-sequence model to generate summaries.
MASS~\cite{mass} is a pre-trained sequence-to-sequence model based on Transformer networks.
Experimental results show that \ours{} achieves better performance than previous work.
Besides, in the low-resource setting (i.e., only 10,000 examples are used as training data), our model outperforms MASS by $7.08$ point in ROUGE-L.

\subsection{Question Answering (QA)}
\label{sec:qa}

The task is to answer a question given a passage~\cite{squad1, squad2,gao2019neural}. There are two settings. The first is called \emph{extractive} QA, where the answer is assumed to be a text span in the passage. The other is called \emph{generative} QA, where the answer needs to be generated on the fly.

%We evaluate the model on reading comprehension style question answering, which answers questions from a given passage~\cite{squad1, squad2}. A recent survey is~\cite{gao2019neural}. Reading comprehension has been mainly tackled with two types of approaches.  
%First, extractive models extract a subspan from the input passage as the answer to the question. Second, generative models use the question and passage as inputs, and generate a free-form answer with a sequence-to-sequence model. We evaluate \ours{} in both settings. 

\paragraph{Extractive QA}
\label{sec:ext:qa}

This task can be formulated as a NLU task where we need to predict the start and end positions of the answer spans within the passage.
We fine-tune the pre-trained \ours{} as a bidirectional encoder for the task.
We conduct experiments on the Stanford Question Answering Dataset (SQuAD) 2.0~\citep{squad2}, and Conversational Question Answering (CoQA)~\citep{coqa} datasets.

%Extractive question answering extracts a continuous span from the given passage to answer the question. The task is usually formulated to predict the start and end positions of the answer spans within the passage.

%We apply our pre-trained LM as a bidirectional encoder for the task. Following~\cite{bert}, we pack the input question and passage into one sequence ``\sptk{SOS} \textit{question} \sptk{EOS} \textit{passage} \sptk{EOS}''. The question is used as the first segment, and the passage is the second one in \ours.

%The final hidden vector of each token is fed into two softmax classifiers to predict the probability of the token being the start or end positions of the answer span. The training objective is to maximize the likelihood of the correct start and end positions.

% We use the same method as in~\cite{bert} to fine-tune the SQuAD 2.0.
The results on SQuAD 2.0 are reported in Table~\ref{tbl:squad:result}, where we compare two models in Exact Match (EM) and F1 score. 
RMR+ELMo~\cite{hu-mrc-2018} is an LSTM-based question answering model augmented with pre-trained language representation.
BERT$_{\text{LARGE}}$ is a cased model, fine-tuned on the SQuAD training data for $3$ epochs, with batch size $24$, and maximum length $384$.
\ours{} is fine-tuned in the same way as BERT$_{\text{LARGE}}$.
We see that \ours{} outperforms BERT$_{\text{LARGE}}$.

CoQA is a conversational question answering dataset.
Compared with SQuAD, CoQA has several unique characteristics. First, the examples in CoQA are conversational, so we need to answer the input question based on conversation histories. Second, the answers in CoQA can be free-form texts, including a large portion is of yes/no answers.

We modify the model used for SQuAD as follows.
Firstly, in addition to the asked question, we concatenate the question-answer histories to the first segment, so that the model can capture conversational information.
Secondly, for yes/no questions, we use the final hidden vector of the \sptk{SOS} token to predict whether the input is a yes/no question, and whether the answer is \texttt{yes} or \texttt{no}. For other examples, we select a passage subspan with the highest F1 score for training. 

The results on CoQA are reported in Table~\ref{tbl:coqa:extractive:result}, where we compare two models in F1 scores. 
DrQA+ELMo~\citep{coqa} is an LSTM-based question answering model augmented with pre-trained ELMo representation.
BERT$_{\text{LARGE}}$ is a cased model, fine-tuned on the CoQA training data for $2$ epochs, with batch size $16$, and maximum length $512$. \ours{} is fine-tuned with the same hyper-parameters as BERT$_{\text{LARGE}}$. 
We see that \ours{} outperforms BERT$_\text{LARGE}$.

\begin{table}[t]
\centering
\small
\begin{minipage}{1.9in}
\centering
\begin{tabular}{l c c c}
\toprule
                & EM       & F1  
                \\ \midrule
RMR+ELMo~\citep{hu-mrc-2018} & 71.4 & 73.7 \\
BERT$_{\text{LARGE}}$ & 78.9           & 81.8  \\
\ours   & \textbf{80.5} & \textbf{83.4}  \\
\bottomrule
\end{tabular}
\caption{Extractive QA results on the SQuAD development set.}
\normalsize
\label{tbl:squad:result}
\end{minipage}
\hfill
\begin{minipage}{1.7in}
\centering
\begin{tabular}{l c c c}
\toprule
                & F1    
                \\ \midrule
DrQA+ELMo~\citep{coqa} & 67.2           \\
BERT$_\text{LARGE}$ & 82.7           \\
\ours   & \textbf{84.9}  \\ \bottomrule
\end{tabular}
\caption{Extractive QA results on the CoQA development set.}
\normalsize
\label{tbl:coqa:extractive:result}
\end{minipage}
\hfill
\begin{minipage}{1.7in}
\centering
\begin{tabular}{l c c c}
\toprule
                & F1    
                \\ \midrule
Seq2Seq~\citep{coqa}   & 27.5      \\
PGNet~\citep{coqa}   & 45.4      \\
\ours & \textbf{82.5}  \\
\bottomrule
\end{tabular}
\caption{Generative QA results on the CoQA development set.}
\normalsize
\label{tbl:coqa:generative:result}
\end{minipage}
\end{table}

\paragraph{Generative QA}

Generative question answering generates free-form answers for the input question and passage, which is a NLG task.
In contrast, extractive methods can only predict subspans of the input passage as answers.
On the CoQA dataset (as described above), \newcite{coqa} show that vanilla sequence-to-sequence models still underperforms extractive methods by a wide margin.

We adapt \ours{} to generative question answering as a sequence-to-sequence model.
The first segment (i.e., the input sequence) is the concatenation of conversational histories, the input question and the passage. The second segment (i.e., the output sequence) is the answer. 
We fine-tune the pre-trained \ours{} on the CoQA training set for $10$ epochs. We set the batch size to $32$, the mask probability to $0.5$, and the maximum length to $512$. We also use label smoothing with rate of $0.1$.
The other hyper-parameters are kept the same as pre-training. 
During decoding, we use beam search with beam size of $3$. The maximum length of input question and passage is $470$. For passages that are longer than the maximum length, we split the passage into several chunks with a sliding window approach, and select a chunk with the highest word overlap over the question.

We compare our method with the generative question answering models Seq2Seq and PGNet as described in~\cite{coqa}.
The Seq2Seq baseline is a sequence-to-sequence model with an attention mechanism. The PGNet model augments Seq2Seq with a copy mechanism.
As shown in Table~\ref{tbl:coqa:generative:result}, our generative question answering model outperforms previous generative methods by a wide margin, which significantly closes the gap between generative method and extractive method.

\subsection{Question Generation}
\label{sec:qg}

We conduct experiments for the answer-aware question generation task~\cite{zhou-qg-2017}. Given an input passage and an answer span, our goal is to generate a question that asks for the answer.
The SQuAD 1.1 dataset~\cite{squad1} is used for evaluation.
Following~\cite{du-qg-2017}, we split the original training set into training and test sets, and keep the original development set.
We also conduct experiments following the data split as in~\cite{zhao-qg-2018}, which uses the reversed dev-test split.

The question generation task is formulated as a sequence-to-sequence problem. The first segment is the concatenation of input passage and answer, while the second segment is the generated question.

% The input passage, the answer, and the generated question are packed together into a sequence ``\sptk{SOS} \textit{passage} \sptk{EOS} \textit{answer} \sptk{EOS} \textit{question} \sptk{EOS}''. Both the input passage and answer are regarded as the first text segment, while the generated question is the second segment in the unified LM. 
We fine-tune \ours{} on the training set for $10$ epochs. We set batch size to $32$, masking probability to $0.7$, and learning rate to 2e-5. The rate of label smoothing is $0.1$. The other hyper-parameters are the same as pre-training.
During decoding, we truncate the input to $464$ tokens by selecting a passage chunk which contains the answer.
The evaluation metrics BLEU-4, METEOR, and ROUGE-L are computed by the same scripts as in~\cite{du-qg-2017}.

The results\footnote{Notice that if we directly use the tokenized references provided by~\newcite{du-qg-2017}, the results are (21.63 BLEU-4 / 25.04 METEOR / 51.09 ROUGE-L) on the original data split~\cite{du-qg-2017}, and (23.08 BLEU-4 / 25.57 METEOR / 52.03 ROUGE-L) in the reversed dev-test setup~\cite{zhao-qg-2018}.} are presented in Table~\ref{tbl:qg:result}.
CorefNQG~\cite{du-qg-2018} is based on a sequence-to-sequence model with attention and a feature-rich encoder.
MP-GSN~\cite{zhao-qg-2018} uses an attention-based sequence-to-sequence model with a gated self-attention encoder.
SemQG~\cite{zhang-qg-2019} uses two semantics-enhanced rewards to regularize the generation.
\ours{} outperforms previous models and achieves a new state-of-the-art for question generation. % on the SQuAD 1.1 dataset.

\begin{table}[t]
\centering
\small
\begin{minipage}{2.8in}
\centering
\begin{tabular}{l c c c}
\toprule
& BLEU-4       & MTR        & RG-L
\\ \midrule
CorefNQG~\cite{du-qg-2018} & 15.16           & 19.12   & - \\
SemQG~\cite{zhang-qg-2019} & 18.37           & 22.65   & 46.68 \\
\ours & \textbf{22.12}  & \textbf{25.06} & \textbf{51.07} \\ \midrule
MP-GSN~\cite{zhao-qg-2018}   & 16.38 & 20.25           &   44.48   \\
SemQG~\cite{zhang-qg-2019} & 20.76           & 24.20   & 48.91 \\
\ours & \textbf{23.75}  & \textbf{25.61} & \textbf{52.04} \\
\bottomrule
\end{tabular}
\caption{Question generation results on SQuAD. MTR is short for METEOR, and RG for ROUGE. Results in the groups use different data splits.}
\normalsize
\label{tbl:qg:result}
\end{minipage}
\hfill
\begin{minipage}{2.6in}
\centering
\begin{tabular}{l c c c}
\toprule
                & EM       & F1  
                \\ \midrule
\ours{} QA Model (Section~\ref{sec:qa})   & 80.5 & 83.4  \\
~~+ \ours{} Generated Questions   & \textbf{84.7} & \textbf{87.6}       \\
\bottomrule
\end{tabular}
\normalsize
\caption{Question generation based on \ours{} improves question answering results on the SQuAD development set.}
\label{tbl:squad:qg4qa:result}
\end{minipage}
\end{table}

\paragraph{Generated Questions Improve QA}

The question generation model can automatically harvest a large number of question-passage-answer examples from a text corpus.
We show that the augmented data generated by question generation improves the question answering model.

We generate five million answerable examples, and four million unanswerable examples by modifying the answerable ones.
We fine-tune our question answering model on the generated data for one epoch. Then the model is fine-tuned on the SQuAD 2.0 data for two more epochs. 

As shown in Table~\ref{tbl:squad:qg4qa:result}, the augmented data generated by \ours{} improves question answering model introduced in Section~\ref{sec:qa}.
Note that we use bidirectional masked language modeling as an auxiliary task for both the generated and SQuAD 2.0 datasets during fine-tuning, which brings $2.3$ absolute improvement compared to directly using automatically generated examples. A possible reason is that the auxiliary task alleviates catastrophic forgetting~\cite{general:ling:intell} when fine-tuning on augmented data.

\subsection{Response Generation}

\begin{table}[t]
\centering
\small
\begin{tabular}{@{ \hskip2pt}l @{ \hskip2pt}l@{ \hskip2pt} @{ \hskip2pt}c@{ \hskip2pt} @{ \hskip2pt}c@{ \hskip2pt} @{ \hskip2pt}c@{ \hskip2pt} @{ \hskip2pt}c@{ \hskip2pt} @{ \hskip2pt}c@{ \hskip2pt} @{ \hskip1pt}c@{ \hskip1pt} @{ \hskip1pt}c @{ \hskip1pt} @{ \hskip1pt} c@{ \hskip1pt}  @{ \hskip2pt} c@{ \hskip2pt} @{ \hskip2pt} c@{ \hskip2pt}}
\toprule
& NIST-4    & BLEU-4 & METEOR & Entropy-4 & Div-1 & Div-2 & Avg len \\ \midrule
Best System in DSTC7 Shared Task & 2.523 & 1.83  & 8.07 & 9.030 & 0.109 & 0.325& 15.133 \\
\ours & \textbf{2.669} & \textbf{4.39} & \textbf{8.27} & \textbf{9.195} & \textbf{0.120} & \textbf{0.391} & 14.807 \\
\midrule
Human Performance & 2.650 & 3.13 & 8.31 & 10.445 & 0.167 & 0.670 & 18.76 \\
\bottomrule
\end{tabular}
\caption{Response generation results. Div-1 and Div-2 indicate diversity of unigrams and bigrams, respectively.}
\label{tbl:dgrg:result}
\end{table}

We evaluate \ours{} on the document-grounded dialog response generation task~\cite{conversing:dataset,gao2019neural}.
Given a multi-turn conversation history and a web document as the knowledge source, the system needs to generate a natural language response that is both conversationally appropriate and reflective of the contents of the web document.
We fine-tune \ours{} to the task as a sequence-to-sequence model. The first segment (input sequence) is the concatenation of the web document and the conversation history. The second segment (output sequence) is the response.
We fine-tune \ours{} on the DSTC7 training data for $20$ epochs, with batch size $64$.
The masking probability is set to $0.5$. The maximum length is $512$.
During decoding, we use beam search with size of $10$. The maximum length of generated response is set to $40$.
As shown in Table~\ref{tbl:dgrg:result}, \ours{} outperforms the best system~\cite{tam2019cluster} in the DSTC7 shared task~\cite{dstc7} across all evaluation metrics.

\subsection{GLUE Benchmark}

We evaluate \ours{} on the General Language Understanding Evaluation (GLUE) benchmark~\cite{wang2018glue}. GLUE is a collection of nine language understanding tasks,
including question answering~\cite{squad1}, linguistic acceptability~\cite{cola2018}, sentiment analysis~\cite{sst2013}, text similarity~\cite{sts-b2017}, paraphrase detection~\cite{mrpc2005}, and natural language inference (NLI)~\cite{rte1,rte2,rte3,rte5,winograd2012,mnli2017}.
% \footnote{The Quora Question Pairs (QQP) dataset is from: \url{https://data.quora.com/First-Quora-Dataset-Release-Question-Pairs}}

Our model is fine-tuned as a bidirectional LM.
We use Adamax~\cite{kingma2014adam} as our optimizer with a learning rate of 5e-5 and a batch size of $32$. The maximum number of epochs is set to $5$. A linear learning rate decay schedule with warmup of $0.1$ is used. The dropout rate of the last linear projection for each task is set to $0.1$, except $0.3$ for MNLI and $0.05$ for CoLA/SST-2. 
To avoid the gradient explosion issue, the gradient norm was clipped within $1$. We truncated the tokens no longer than $512$.

\begin{table*}[t]
\centering
\small
\begin{tabular}{@{ \hskip2pt}l @{ \hskip2pt}l@{ \hskip2pt} @{ \hskip2pt}c@{ \hskip2pt} @{ \hskip2pt}c@{ \hskip2pt} @{ \hskip2pt}c@{ \hskip2pt} @{ \hskip2pt}c@{ \hskip2pt} @{ \hskip2pt}c@{ \hskip2pt} @{ \hskip1pt}c@{ \hskip1pt} @{ \hskip1pt}c @{ \hskip1pt} @{ \hskip1pt} c@{ \hskip1pt}  @{ \hskip2pt} c@{ \hskip2pt} @{ \hskip2pt} c@{ \hskip2pt}}
\toprule
\textbf{Model} &CoLA&	SST-2 &MRPC& STS-B&QQP&MNLI-m/mm&QNLI&RTE&WNLI&AX &\textbf{Score}\\ 
% & 8.5k &67k &3.7k &7k &364k &393k &108k &2.5k &634 & & \\ 
 &MCC &Acc &F1&S Corr&F1 &Acc &Acc &Acc &Acc &Acc & \\  \midrule
% BiLSTM+ELMo+Attn $^1$&36.0 &90.4 &84.9/77.9 &75.1/73.3 &64.8/84.7 &76.4/76.1 &79.8 &56.8 &65.1 &26.5 &70.0 \\ \midrule
GPT &45.4 &91.3 &82.3&80.0 &70.3 &82.1/81.4 &87.4 &56.0 &53.4  &29.8 &72.8 \\
BERT$_{\text{LARGE}}$ & 60.5 &\textbf{94.9} &89.3 &86.5 &\textbf{72.1} &86.7/\textbf{85.9} & \textbf{92.7} & 70.1 &65.1	&39.6 & 80.5\\ 
\ours{} & \textbf{61.1} &94.5 &\textbf{90.0} &\textbf{87.7} &71.7 &\textbf{87.0/85.9} &\textbf{92.7} &\textbf{70.9} &65.1	&38.4 & \textbf{80.8}\\ \bottomrule
% {Human} &66.4&97.8&86.3    &92.6	&59.5	&92.0/92.8	&91.2	&93.6	&95.9 &- & 87.1\\ \bottomrule
\end{tabular}
\caption{GLUE test set results scored using the GLUE evaluation server.}
\label{tbl:glue_test}
\end{table*}

Table~\ref{tbl:glue_test} presents the GLUE test results obtained from the benchmark evaluation server. The results show that \ours{} obtains comparable performance on the GLUE tasks in comparison with BERT$_{\text{LARGE}}$.

\section{Conclusion and Future Work}

We propose a unified pre-training model, \ours{}, which is jointly optimized for several LM objectives with shared parameters. The unification of bidirectional, unidirectional, and sequence-to-sequence LMs enables us to straightforwardly fine-tune the pre-trained \ours{} for both NLU and NLG tasks. Experimental results demonstrate that our model compares favorably with BERT on the GLUE benchmark and two question answering datasets. 
In addition, \ours{} outperforms previous state-of-the-art models on five NLG datasets: CNN/DailyMail and Gigaword abstractive summarization, SQuAD question generation, CoQA generative question answering, and DSTC7 dialog response generation.

The work can be advanced from the following perspectives:
\begin{itemize}[leftmargin=*]
\setlength\itemsep{0.01em}
\item We will push the limit of the current method by training more epochs and larger models on web-scale text corpora. At the same time, we will also conduct more experiments on end applications as well as ablation experiments to investigate the model capability and the benefits of pre-training multiple language modeling tasks with the same network.
\item We are focusing on monolingual NLP tasks in our current experiments. We are also interested in extending \ours{} to support cross-lingual tasks~\cite{xnlg}.
\item We will conduct multi-task fine-tuning on both NLU and NLG tasks, which is a natural extension of Multi-Task Deep Neural Network (MT-DNN)~\cite{mt-dnn}.
% to support both language understanding and language generation tasks.
%\item We are also interested in extending our work to multi-modality settings where image or video information can be effectively leveraged in the LM pre-training in addition to pure text information as used in this work.
\end{itemize}

\paragraph{Acknowledgement}
We would like to acknowledge Shiyue Zhang for the helpful discussions about the question generation experiments.

\bibliographystyle{plainnat}
\bibliography{unified}

\begin{thebibliography}{53}
\providecommand{\natexlab}[1]{#1}
\providecommand{\url}[1]{\texttt{#1}}
\expandafter\ifx\csname urlstyle\endcsname\relax
  \providecommand{\doi}[1]{doi: #1}\else
  \providecommand{\doi}{doi: \begingroup \urlstyle{rm}\Url}\fi

\bibitem[Baevski et~al.(2019)Baevski, Edunov, Liu, Zettlemoyer, and
  Auli]{clozepretrain19}
Alexei Baevski, Sergey Edunov, Yinhan Liu, Luke Zettlemoyer, and Michael Auli.
\newblock Cloze-driven pretraining of self-attention networks.
\newblock \emph{arXiv preprint arXiv:1903.07785}, 2019.

\bibitem[Bar-Haim et~al.(2006)Bar-Haim, Dagan, Dolan, Ferro, and
  Giampiccolo]{rte2}
Roy Bar-Haim, Ido Dagan, Bill Dolan, Lisa Ferro, and Danilo Giampiccolo.
\newblock The second {PASCAL} recognising textual entailment challenge.
\newblock In \emph{Proceedings of the Second {PASCAL} Challenges Workshop on
  Recognising Textual Entailment}, 01 2006.

\bibitem[Bentivogli et~al.(2009)Bentivogli, Dagan, Dang, Giampiccolo, and
  Magnini]{rte5}
Luisa Bentivogli, Ido Dagan, Hoa~Trang Dang, Danilo Giampiccolo, and Bernardo
  Magnini.
\newblock The fifth pascal recognizing textual entailment challenge.
\newblock In \emph{In Proc Text Analysis Conference (TAC-09)}, 2009.

\bibitem[Cao et~al.(2018)Cao, Li, Li, and Wei]{re3sum}
Ziqiang Cao, Wenjie Li, Sujian Li, and Furu Wei.
\newblock Retrieve, rerank and rewrite: Soft template based neural
  summarization.
\newblock In \emph{Proceedings of the 56th Annual Meeting of the Association
  for Computational Linguistics}, pages 152--161, Melbourne, Australia, July
  2018. Association for Computational Linguistics.

\bibitem[Cer et~al.(2017)Cer, Diab, Agirre, Lopez-Gazpio, and
  Specia]{sts-b2017}
Daniel Cer, Mona Diab, Eneko Agirre, Inigo Lopez-Gazpio, and Lucia Specia.
\newblock Semeval-2017 task 1: Semantic textual similarity-multilingual and
  cross-lingual focused evaluation.
\newblock \emph{arXiv preprint arXiv:1708.00055}, 2017.

\bibitem[Chi et~al.(2019)Chi, Dong, Wei, Wang, Mao, and Huang]{xnlg}
Zewen Chi, Li~Dong, Furu Wei, Wenhui Wang, Xian-Ling Mao, and Heyan Huang.
\newblock Cross-lingual natural language generation via pre-training.
\newblock \emph{ArXiv}, abs/1909.10481, 2019.

\bibitem[Dagan et~al.(2006)Dagan, Glickman, and Magnini]{rte1}
Ido Dagan, Oren Glickman, and Bernardo Magnini.
\newblock The pascal recognising textual entailment challenge.
\newblock In \emph{Proceedings of the First International Conference on Machine
  Learning Challenges: Evaluating Predictive Uncertainty Visual Object
  Classification, and Recognizing Textual Entailment}, MLCW'05, pages 177--190,
  Berlin, Heidelberg, 2006. Springer-Verlag.

\bibitem[Dai and Le(2015)]{semi_seq}
Andrew~M Dai and Quoc~V Le.
\newblock Semi-supervised sequence learning.
\newblock In \emph{Advances in Neural Information Processing Systems 28}, pages
  3079--3087. Curran Associates, Inc., 2015.

\bibitem[Devlin et~al.(2018)Devlin, Chang, Lee, and Toutanova]{bert}
Jacob Devlin, Ming{-}Wei Chang, Kenton Lee, and Kristina Toutanova.
\newblock {BERT:} pre-training of deep bidirectional transformers for language
  understanding.
\newblock \emph{CoRR}, abs/1810.04805, 2018.

\bibitem[Dolan and Brockett(2005)]{mrpc2005}
William~B Dolan and Chris Brockett.
\newblock Automatically constructing a corpus of sentential paraphrases.
\newblock In \emph{Proceedings of the Third International Workshop on
  Paraphrasing (IWP2005)}, 2005.

\bibitem[Du and Cardie(2018)]{du-qg-2018}
Xinya Du and Claire Cardie.
\newblock Harvesting paragraph-level question-answer pairs from {W}ikipedia.
\newblock In \emph{Proceedings of the 56th Annual Meeting of the Association
  for Computational Linguistics}, pages 1907--1917, Melbourne, Australia, July
  2018. Association for Computational Linguistics.

\bibitem[Du et~al.(2017)Du, Shao, and Cardie]{du-qg-2017}
Xinya Du, Junru Shao, and Claire Cardie.
\newblock Learning to ask: Neural question generation for reading
  comprehension.
\newblock In \emph{Proceedings of the 55th Annual Meeting of the Association
  for Computational Linguistics, {ACL} 2017, Vancouver, Canada, July 30 -
  August 4, Volume 1: Long Papers}, pages 1342--1352, 2017.

\bibitem[Edunov et~al.(2019)Edunov, Baevski, and Auli]{Edunov2019PretrainedLM}
Sergey Edunov, Alexei Baevski, and Michael Auli.
\newblock Pre-trained language model representations for language generation.
\newblock \emph{CoRR}, abs/1903.09722, 2019.

\bibitem[Galley et~al.(2019)Galley, Brockett, Gao, Gao, and Dolan]{dstc7}
Michel Galley, Chris Brockett, Xiang Gao, Jianfeng Gao, and Bill Dolan.
\newblock Grounded response generation task at dstc7.
\newblock In \emph{AAAI Dialog System Technology Challenges Workshop}, 2019.

\bibitem[Gao et~al.(2019)Gao, Galley, Li, et~al.]{gao2019neural}
Jianfeng Gao, Michel Galley, Lihong Li, et~al.
\newblock Neural approaches to conversational ai.
\newblock \emph{Foundations and Trends in Information Retrieval}, 13\penalty0
  (2-3):\penalty0 127--298, 2019.

\bibitem[Gehrmann et~al.(2018)Gehrmann, Deng, and
  Rush]{gehrmann-etal-2018-bottom}
Sebastian Gehrmann, Yuntian Deng, and Alexander Rush.
\newblock Bottom-up abstractive summarization.
\newblock In \emph{Proceedings of the 2018 Conference on Empirical Methods in
  Natural Language Processing}, pages 4098--4109, Brussels, Belgium,
  October-November 2018. Association for Computational Linguistics.

\bibitem[Giampiccolo et~al.(2007)Giampiccolo, Magnini, Dagan, and Dolan]{rte3}
Danilo Giampiccolo, Bernardo Magnini, Ido Dagan, and Bill Dolan.
\newblock The third {PASCAL} recognizing textual entailment challenge.
\newblock In \emph{Proceedings of the {ACL}-{PASCAL} Workshop on Textual
  Entailment and Paraphrasing}, pages 1--9, Prague, June 2007. Association for
  Computational Linguistics.

\bibitem[Hendrycks and Gimpel(2016)]{gelu}
Dan Hendrycks and Kevin Gimpel.
\newblock Gaussian error linear units ({GELU}s).
\newblock \emph{arXiv preprint arXiv:1606.08415}, 2016.

\bibitem[Howard and Ruder(2018)]{ulmfit}
Jeremy Howard and Sebastian Ruder.
\newblock Universal language model fine-tuning for text classification.
\newblock In \emph{Proceedings of the 56th Annual Meeting of the Association
  for Computational Linguistics}, pages 328--339, Melbourne, Australia, July
  2018. Association for Computational Linguistics.

\bibitem[Hu et~al.(2018)Hu, Wei, Peng, Huang, Yang, and Zhou]{hu-mrc-2018}
Minghao Hu, Furu Wei, Yuxing Peng, Zhen Huang, Nan Yang, and Ming Zhou.
\newblock Read + verify: Machine reading comprehension with unanswerable
  questions.
\newblock \emph{CoRR}, abs/1808.05759, 2018.

\bibitem[Kingma and Ba(2014)]{kingma2014adam}
Diederik Kingma and Jimmy Ba.
\newblock Adam: A method for stochastic optimization.
\newblock \emph{arXiv preprint arXiv:1412.6980}, 2014.

\bibitem[Kingma and Ba(2015)]{adam}
Diederik~P. Kingma and Jimmy Ba.
\newblock Adam: {A} method for stochastic optimization.
\newblock In \emph{3rd International Conference on Learning Representations},
  San Diego, CA, 2015.

\bibitem[Klein et~al.(2017)Klein, Kim, Deng, Senellart, and Rush]{opennmt}
Guillaume Klein, Yoon Kim, Yuntian Deng, Jean Senellart, and Alexander Rush.
\newblock {O}pen{NMT}: Open-source toolkit for neural machine translation.
\newblock In \emph{Proceedings of {ACL} 2017, System Demonstrations}, pages
  67--72, Vancouver, Canada, July 2017. Association for Computational
  Linguistics.

\bibitem[Levesque et~al.(2012)Levesque, Davis, and Morgenstern]{winograd2012}
Hector Levesque, Ernest Davis, and Leora Morgenstern.
\newblock The winograd schema challenge.
\newblock In \emph{Thirteenth International Conference on the Principles of
  Knowledge Representation and Reasoning}, 2012.

\bibitem[Lin(2004)]{lin-2004-rouge}
Chin-Yew Lin.
\newblock {ROUGE}: A package for automatic evaluation of summaries.
\newblock In \emph{Text Summarization Branches Out: Proceedings of the {ACL}-04
  Workshop}, pages 74--81, Barcelona, Spain, July 2004. Association for
  Computational Linguistics.

\bibitem[Liu et~al.(2019)Liu, He, Chen, and Gao]{mt-dnn}
Xiaodong Liu, Pengcheng He, Weizhu Chen, and Jianfeng Gao.
\newblock Multi-task deep neural networks for natural language understanding.
\newblock \emph{CoRR}, abs/1901.11504, 2019.

\bibitem[Liu(2019)]{bertsum}
Yang Liu.
\newblock Fine-tune {BERT} for extractive summarization.
\newblock \emph{CoRR}, abs/1903.10318, 2019.

\bibitem[Paulus et~al.(2018)Paulus, Xiong, and Socher]{Paulus2018ADR}
Romain Paulus, Caiming Xiong, and Richard Socher.
\newblock A deep reinforced model for abstractive summarization.
\newblock \emph{CoRR}, abs/1705.04304, 2018.

\bibitem[Peters et~al.(2018)Peters, Neumann, Iyyer, Gardner, Clark, Lee, and
  Zettlemoyer]{elmo}
Matthew Peters, Mark Neumann, Mohit Iyyer, Matt Gardner, Christopher Clark,
  Kenton Lee, and Luke Zettlemoyer.
\newblock Deep contextualized word representations.
\newblock In \emph{Proceedings of the 2018 Conference of the North American
  Chapter of the Association for Computational Linguistics: Human Language
  Technologies}, pages 2227--2237, New Orleans, Louisiana, June 2018.
  Association for Computational Linguistics.

\bibitem[Qin et~al.(2019)Qin, Galley, Brockett, Liu, Gao, Dolan, Choi, and
  Gao]{conversing:dataset}
Lianhui Qin, Michel Galley, Chris Brockett, Xiaodong Liu, Xiang Gao, Bill
  Dolan, Yejin Choi, and Jianfeng Gao.
\newblock Conversing by reading: Contentful neural conversation with on-demand
  machine reading.
\newblock In \emph{Proceedings of the 57th Annual Meeting of the Association
  for Computational Linguistics}, pages 5427--5436, Florence, Italy, July 2019.
  Association for Computational Linguistics.

\bibitem[Radford et~al.(2018)Radford, Narasimhan, Salimans, and Sutskever]{gpt}
Alec Radford, Karthik Narasimhan, Tim Salimans, and Ilya Sutskever.
\newblock Improving language understanding by generative pre-training.
\newblock 2018.

\bibitem[Radford et~al.(2019)Radford, Wu, Child, Luan, Amodei, and
  Sutskever]{gpt2}
Alec Radford, Jeff Wu, Rewon Child, David Luan, Dario Amodei, and Ilya
  Sutskever.
\newblock Language models are unsupervised multitask learners.
\newblock 2019.

\bibitem[Rajpurkar et~al.(2016)Rajpurkar, Zhang, Lopyrev, and Liang]{squad1}
Pranav Rajpurkar, Jian Zhang, Konstantin Lopyrev, and Percy Liang.
\newblock {SQ}u{AD}: 100,000+ questions for machine comprehension of text.
\newblock In \emph{Proceedings of the 2016 Conference on Empirical Methods in
  Natural Language Processing}, pages 2383--2392, Austin, Texas, November 2016.
  Association for Computational Linguistics.

\bibitem[Rajpurkar et~al.(2018)Rajpurkar, Jia, and Liang]{squad2}
Pranav Rajpurkar, Robin Jia, and Percy Liang.
\newblock Know what you don't know: Unanswerable questions for {SQuAD}.
\newblock In \emph{Proceedings of the 56th Annual Meeting of the Association
  for Computational Linguistics, {ACL} 2018, Melbourne, Australia, July 15-20,
  2018, Volume 2: Short Papers}, pages 784--789, 2018.

\bibitem[Reddy et~al.(2019)Reddy, Chen, and Manning]{coqa}
Siva Reddy, Danqi Chen, and Christopher~D. Manning.
\newblock {C}o{QA}: A conversational question answering challenge.
\newblock \emph{Transactions of the Association for Computational Linguistics},
  7:\penalty0 249--266, March 2019.

\bibitem[Rush et~al.(2015)Rush, Chopra, and Weston]{abs}
Alexander~M. Rush, Sumit Chopra, and Jason Weston.
\newblock A neural attention model for abstractive sentence summarization.
\newblock In \emph{Proceedings of the 2015 Conference on Empirical Methods in
  Natural Language Processing}, pages 379--389, Lisbon, Portugal, September
  2015. Association for Computational Linguistics.

\bibitem[See et~al.(2017)See, Liu, and Manning]{see-etal-2017-get}
Abigail See, Peter~J. Liu, and Christopher~D. Manning.
\newblock Get to the point: Summarization with pointer-generator networks.
\newblock In \emph{Proceedings of the 55th Annual Meeting of the Association
  for Computational Linguistics}, pages 1073--1083, Vancouver, Canada, July
  2017. Association for Computational Linguistics.

\bibitem[Socher et~al.(2013)Socher, Perelygin, Wu, Chuang, Manning, Ng, and
  Potts]{sst2013}
Richard Socher, Alex Perelygin, Jean Wu, Jason Chuang, Christopher~D Manning,
  Andrew Ng, and Christopher Potts.
\newblock Recursive deep models for semantic compositionality over a sentiment
  treebank.
\newblock In \emph{Proceedings of the 2013 conference on empirical methods in
  natural language processing}, pages 1631--1642, 2013.

\bibitem[Song et~al.(2019)Song, Tan, Qin, Lu, and Liu]{mass}
Kaitao Song, Xu~Tan, Tao Qin, Jianfeng Lu, and Tie-Yan Liu.
\newblock Mass: Masked sequence to sequence pre-training for language
  generation.
\newblock \emph{arXiv preprint arXiv:1905.02450}, 2019.

\bibitem[Szegedy et~al.(2016)Szegedy, Vanhoucke, Ioffe, Shlens, and
  Wojna]{label:smoothing}
Christian Szegedy, Vincent Vanhoucke, Sergey Ioffe, Jon Shlens, and Zbigniew
  Wojna.
\newblock Rethinking the inception architecture for computer vision.
\newblock In \emph{Proceedings of the IEEE Conference on Computer Vision and
  Pattern Recognition}, pages 2818--2826, 2016.

\bibitem[Tam et~al.(2019)Tam, Ding, Niu, and Zhou]{tam2019cluster}
Y~Tam, Jiachen Ding, Cheng Niu, and Jie Zhou.
\newblock Cluster-based beam search for pointer-generator chatbot grounded by
  knowledge.
\newblock In \emph{AAAI Dialog System Technology Challenges Workshop}, 2019.

\bibitem[Taylor(1953)]{taylor1953cloze}
Wilson~L Taylor.
\newblock Cloze procedure: A new tool for measuring readability.
\newblock \emph{Journalism Bulletin}, 30\penalty0 (4):\penalty0 415--433, 1953.

\bibitem[Vaswani et~al.(2017)Vaswani, Shazeer, Parmar, Uszkoreit, Jones, Gomez,
  Kaiser, and Polosukhin]{transformer}
Ashish Vaswani, Noam Shazeer, Niki Parmar, Jakob Uszkoreit, Llion Jones,
  Aidan~N Gomez, {\L}ukasz Kaiser, and Illia Polosukhin.
\newblock Attention is all you need.
\newblock In \emph{Advances in Neural Information Processing Systems 30}, pages
  5998--6008. Curran Associates, Inc., 2017.

\bibitem[Wang and Cho(2019)]{bert_mouth}
Alex Wang and Kyunghyun Cho.
\newblock {BERT} has a mouth, and it must speak: {BERT} as a markov random
  field language model.
\newblock \emph{CoRR}, abs/1902.04094, 2019.

\bibitem[Wang et~al.(2019)Wang, Singh, Michael, Hill, Levy, and
  Bowman]{wang2018glue}
Alex Wang, Amanpreet Singh, Julian Michael, Felix Hill, Omer Levy, and
  Samuel~R. Bowman.
\newblock {GLUE}: A multi-task benchmark and analysis platform for natural
  language understanding.
\newblock In \emph{International Conference on Learning Representations}, 2019.

\bibitem[Warstadt et~al.(2018)Warstadt, Singh, and Bowman]{cola2018}
Alex Warstadt, Amanpreet Singh, and Samuel~R Bowman.
\newblock Neural network acceptability judgments.
\newblock \emph{arXiv preprint arXiv:1805.12471}, 2018.

\bibitem[Williams et~al.(2018)Williams, Nangia, and Bowman]{mnli2017}
Adina Williams, Nikita Nangia, and Samuel Bowman.
\newblock A broad-coverage challenge corpus for sentence understanding through
  inference.
\newblock In \emph{Proceedings of the 2018 Conference of the North {A}merican
  Chapter of the Association for Computational Linguistics: Human Language
  Technologies}, pages 1112--1122, New Orleans, Louisiana, June 2018.
  Association for Computational Linguistics.

\bibitem[Wu et~al.(2016)Wu, Schuster, Chen, Le, Norouzi, Macherey, Krikun, Cao,
  Gao, Macherey, Klingner, Shah, Johnson, Liu, Kaiser, Gouws, Kato, Kudo,
  Kazawa, Stevens, Kurian, Patil, Wang, Young, Smith, Riesa, Rudnick, Vinyals,
  Corrado, Hughes, and Dean]{gnmt}
Yonghui Wu, Mike Schuster, Zhifeng Chen, Quoc~V. Le, Mohammad Norouzi, Wolfgang
  Macherey, Maxim Krikun, Yuan Cao, Qin Gao, Klaus Macherey, Jeff Klingner,
  Apurva Shah, Melvin Johnson, Xiaobing Liu, Lukasz Kaiser, Stephan Gouws,
  Yoshikiyo Kato, Taku Kudo, Hideto Kazawa, Keith Stevens, George Kurian,
  Nishant Patil, Wei Wang, Cliff Young, Jason Smith, Jason Riesa, Alex Rudnick,
  Oriol Vinyals, Greg Corrado, Macduff Hughes, and Jeffrey Dean.
\newblock Google's neural machine translation system: Bridging the gap between
  human and machine translation.
\newblock \emph{CoRR}, abs/1609.08144, 2016.

\bibitem[Yogatama et~al.(2019)Yogatama, d'Autume, Connor, Kocisky, Chrzanowski,
  Kong, Lazaridou, Ling, Yu, Dyer, and Blunsom]{general:ling:intell}
Dani Yogatama, Cyprien de~Masson d'Autume, Jerome Connor, Tomas Kocisky, Mike
  Chrzanowski, Lingpeng Kong, Angeliki Lazaridou, Wang Ling, Lei Yu, Chris
  Dyer, and Phil Blunsom.
\newblock Learning and evaluating general linguistic intelligence.
\newblock \emph{arXiv preprint arXiv:1901.11373}, 2019.

\bibitem[Zhang and Bansal(2019)]{zhang-qg-2019}
Shiyue Zhang and Mohit Bansal.
\newblock Addressing semantic drift in question generation for semi-supervised
  question answering.
\newblock \emph{CoRR}, abs/1909.06356, 2019.

\bibitem[Zhao et~al.(2018)Zhao, Ni, Ding, and Ke]{zhao-qg-2018}
Yao Zhao, Xiaochuan Ni, Yuanyuan Ding, and Qifa Ke.
\newblock Paragraph-level neural question generation with maxout pointer and
  gated self-attention networks.
\newblock In \emph{Proceedings of the 2018 Conference on Empirical Methods in
  Natural Language Processing}, pages 3901--3910, Brussels, Belgium,
  October-November 2018. Association for Computational Linguistics.

\bibitem[Zhou et~al.(2018)Zhou, Yang, Wei, Tan, Bao, and Zhou]{zhou-qg-2017}
Qingyu Zhou, Nan Yang, Furu Wei, Chuanqi Tan, Hangbo Bao, and Ming Zhou.
\newblock Neural question generation from text: A preliminary study.
\newblock In Xuanjing Huang, Jing Jiang, Dongyan Zhao, Yansong Feng, and
  Yu~Hong, editors, \emph{Natural Language Processing and Chinese Computing},
  pages 662--671. Springer International Publishing, 2018.

\bibitem[Zhu et~al.(2015)Zhu, Kiros, Zemel, Salakhutdinov, Urtasun, Torralba,
  and Fidler]{bookcorpus}
Yukun Zhu, Ryan Kiros, Rich Zemel, Ruslan Salakhutdinov, Raquel Urtasun,
  Antonio Torralba, and Sanja Fidler.
\newblock Aligning books and movies: Towards story-like visual explanations by
  watching movies and reading books.
\newblock In \emph{Proceedings of the IEEE International Conference on Computer
  Vision}, pages 19--27, 2015.

\end{thebibliography}

\begin{appendices}

\section{Long Text Generation: A Case Study}

Our model can generate text samples using the left-to-right setting. We picked three text samples sampled from left to right using our model, as shown in Table~\ref{text-sample}. We use the top-$40$ truncating sampling strategy~\cite{gpt2}, and forbid duplicate 4-grams during generation. For each example, we sampled $10$ times from the same input and we hand-picked the best one; as such, these samples should be considered to be better than the average model output. From the examples, we find that the model can produce fluent output with somewhat consistent contents which fits the inputs' genres and topics. In the first example, given a modified excerpt from the novel ``\textit{1984}'' as input, the model's output is like a short paragraph in an fantasy novel; for the other two with input from Wikipedia, the model makes up ``facts'' with plausible names, time stamps and events written in the Wikipedia style.

\begin{table*}[t]
\begin{center}
\small
\begin{tabular}{p{0.2\linewidth}|p{0.6\linewidth}}
\toprule
\multicolumn{1}{l}{\textbf{Input}} &\multicolumn{1}{c}{\textbf{Output}}
\\ \midrule
Winston sat back. A sense of complete helplessness had descended upon him. & \textit{Winston sat back. A sense of complete helplessness had descended upon him.} He was now helpless and unmoving. What would that mean? To become a monk under her mother's rule. What would it mean for him to be put at the center of the most dangerous of all of demons, the demon known as Azrael? He would never know exactly what that was, or even how it would end, unless he could sense the strength of Azrael's mind and soul. It wasn't only the existence of his mother, it was all he knew, and every moment. He was trapped within her embrace, a trapped, torn, battered, crippling presence he only knew it would mean was eternity. And in the instant that he was lost his life changed forever, when Azraelis' spirit touched his mind, he was engulfed in this dark and mystical, so enchanting and beautiful force. (...)
\\ \midrule
Bert is a yellow Muppet character on the long running children's television show, Sesame Street. &
\textit{Bert is a yellow Muppet character on the long running children's television show, Sesame Street.} He would typically appear as a sidekick or as a villain in the show; he is usually portrayed as a cowardly and untrustworthy figure that tends to intrude on the normal life of the street, such as ``The Story of the Mole'' , ``The Story from the Side'', and ``Bump the Bush''. His voice was provided by Michael Combs. Bert was a regular cast member of Sesame Street, and his appearance was always played throughout the series by two of his regular cast members, Frank Laughner and Michael Combs. (...)
\\ \midrule
Maurice Ravel (7 March 1875 - 28 December 1937) was a French composer, pianist and conductor. & \textit{Maurice Ravel (7 March 1875 - 28 December 1937) was a French composer, pianist and conductor.} He was born in Paris, and studied piano with Eugene Massenet. In the early 1900s Ravel became interested in the composition of music, and he composed much of his early music for the piano. Ravel's early work is marked by a marked sense of lyricism, but in the later years of his career he developed a strong preference for orchestral forms. His first work, ``le Grand Cours d' Auvergne'', was composed in 1909 by Maurice Ravel, a student of Jules Massenet, and was published by A.F.A. in 1912. It was re-published in 1912 by the publisher J.S.D.M. de l'Etablissement Musicale de la Musique Francaise. Ravel wrote the piano concerto ``la Tragedie et la Chanson Dans le Theatre des Champs Elysees'' in 1916. (...) \\ \bottomrule
\end{tabular}
\end{center}
\caption{Text samples generated by our model using left-to-right generation.}
\label{text-sample}
\end{table*}

\section{GLUE Benchmark}

As shown in Table~\ref{tbl:glue:datasets}, we summarize the data size and the evaluation metrics used for the General Language Understanding Evaluation (GLUE) benchmark.

\begin{table}[t]
\centering
\small
\begin{tabular}{@{\hskip3pt}l@{\hskip3pt} @{\hskip3pt}l@{\hskip3pt} @{\hskip3pt}l@{\hskip3pt}}
\toprule \textbf{Corpus} & \textbf{\#Train/\#Dev/\#Test}   &\textbf{Metrics}\\ \midrule
\multicolumn{3}{l}{\emph{Single-Sentence Classification}} \\
CoLA (Acceptability)&8.5k/1k/1k & Matthews corr\\
SST-2 (Sentiment)&67k/872/1.8k & Accuracy\\ \midrule
\multicolumn{3}{l}{\emph{Pairwise Text Classification}} \\
MNLI (NLI)& 393k/20k/20k& Accuracy\\
RTE (NLI) &2.5k/276/3k & Accuracy \\ 
QNLI (NLI)& 108k/5.7k/5.7k& Accuracy\\
WNLI (NLI) &634/71/146& Accuracy \\ 
QQP (Paraphrase)&364k/40k/391k& F1 score\\ 
MRPC (Paraphrase) &3.7k/408/1.7k&F1 score\\ \midrule
\multicolumn{3}{l}{\emph{Text Similarity}} \\
STS-B (Similarity) &7k/1.5k/1.4k & Spearman corr\\ \bottomrule
\end{tabular}
\caption{Summary of the GLUE benchmark.
}
\label{tbl:glue:datasets}
\end{table}

\end{appendices}

\end{document}